%% file: main.tex
\documentclass[conference]{IEEEtran}
\input{notations}

\usepackage{geometry}
\usepackage{pifont}
\usepackage{algorithm}
\usepackage{algorithmic}

\title{Certified ML Object Detection for Surveillance Missions}

\author{
    \IEEEauthorblockN{
    Mohammed Belcaid\IEEEauthorrefmark{1}, 
    Eric Bonnafous\IEEEauthorrefmark{1}, 
    Louis Crison\IEEEauthorrefmark{1}
    Christophe Fauré\IEEEauthorrefmark{1},
    Eric Jenn\IEEEauthorrefmark{2},
    Claire Pagetti\IEEEauthorrefmark{3}
    }
    \IEEEauthorblockA{\IEEEauthorrefmark{1} CS Group, Toulouse, France 
    \IEEEauthorrefmark{2} IRT Saint-Exupéry, Toulouse, France 
    \IEEEauthorrefmark{3} ONERA, Toulouse, France
    }
}

\begin{document}

\maketitle

\section*{Abstract}
In this paper, we present a development process of a drone detection system involving a machine learning object detection component. 
The purpose is to reach acceptable performance objectives and provide sufficient evidences, required by the recommendations (soon to be published) of the ED~324 / ARP~6983 standard, to gain confidence in the dependability of the designed system.

\input{1-introduction}

\input{2-ODD}

\input{3-Dataset}
\input{4-model}

\input{5-deployment}
\input{6-conclusion}

{\footnotesize
\section*{Acknowledgements}
This work has benefited from  
1) the AI Interdisciplinary Institute ANITI,
funded by the “Investing for the Future – PIA3” program Grant agreement
ANR-19-P3IA-0004;
2) the PHYDIAS 2 project funded by the French
government through the France Relance and the NextGenerationEU programs;
3) ARCHEOCS project funded by the 
French Research Agency (ANR) and the partners of the IRT Saint-Exupéry 
Scientific Cooperation Foundation.
}

\bibliographystyle{abbrv}
\bibliography{bib}

\end{document}

%% file: notations.tex
\usepackage[utf8]{inputenc}  
\usepackage[english]{babel}                             
\usepackage{syntax}
\usepackage{tikz}
\usepackage{soul,hyperref}
\usepackage{xcolor,xspace}
\usepackage{listings}
\usepackage[inline]{enumitem}
\usepackage{amsfonts}
\usepackage{amsmath}
\usepackage[font=small,skip=0pt]{caption}
\usepackage{graphicx} 
\usepackage{enumitem}
\setitemize{itemsep=-1pt}

\newtheorem{scenario}{\textbf{Operational scenario}}
\newtheorem{requirement}{\textbf{MLC Requirements}}

%% file: 1-introduction.tex
\section{Introduction}

The ever increasing traffic of UAVs (Unmaned Aerial Vehicles) in the airspace represents a new threat for safety and security. In this context, we are developing a surveillance system aimed at detecting and localizing intrusions of UAVs in sensitive areas. 

\subsection{System description}
The drone surveillance system is composed of two main parts: a \textit{sensing sub-system} and a \textit{machine learning (ML)-based detection and localization sub-system}.
The sensing sub-system is composed of a radar and a camera. The radar scans continuously the area under surveillance and can detect and classify (as \textit{UAV}, \textit{bird} or \textit{other}) multiple objects simultaneously within a range of 5 km and an angle of view of 120 degrees. For small objects, the performance of the radar detection and classification being low ~\cite{interference}, the camera is used to confirm the type of the detected object on the basis of the objects locations provided by the radar.

In this work, we only consider the camera-based detection and localization functions. Therefore, we make no hypothesis on the position of detected objects in the images. This is coherent with the uncertainty inherent to the information provided by the radar. Indeed, assuming the drone to be in specific position in the image (e.g. always in the center), thanks to the radar localization, would possibly lead to miss the presence of an intruder.

\begin{figure}[hbtp]
    \centering
    \includegraphics[scale=0.4]{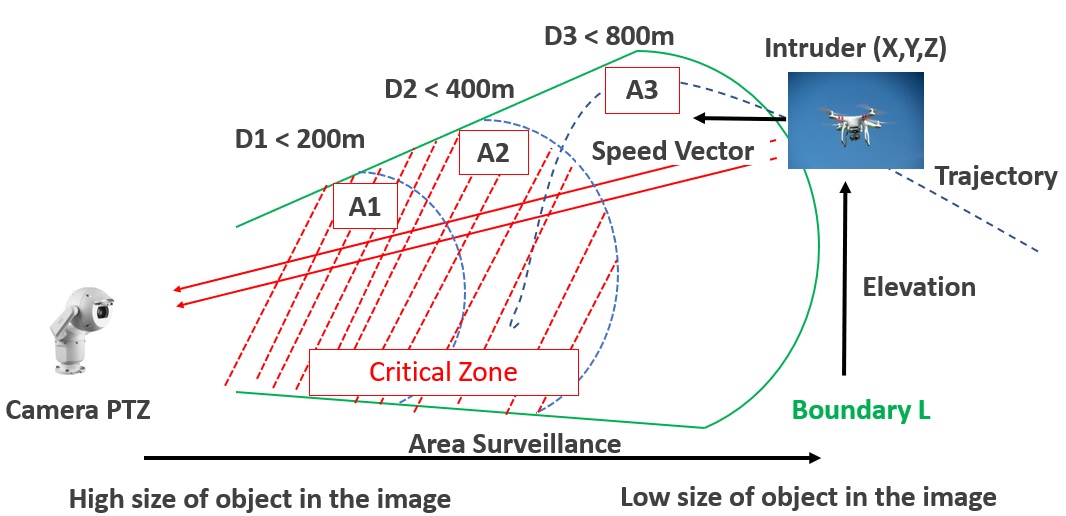}
    \caption{Surveillance Area\label{fig:Surveillance Area}}
\end{figure}  

Figure \ref{fig:Surveillance Area} shows the surveillance area that is delimited by a conic boundary (in green on the figure). In this area, any object of at least 0.5m and at most 800m must be detected. This area is partitioned into three sub-areas (A1, A2, and A3) in order to adapt the detection performance and latency requirements of the system to the distance to the intruder.
Indeed, the closer the intruder, the faster the detection should be and the higher the quality of the detection should be.
Moreover, depending on the size of the drone in the image (in pixel$^2$), the system could execute different object detection models, the performance of which has been \emph{optimized} with a range of object size.

For confidentiality reasons, no precise performance requirements can be given for the system. For area A3 for instance, detection performance must be higher than 80\% and detection latency must be lower than 50ms. In addition, detection performance must be achieved in a large range of environmental conditions including various \emph{backgrounds} (landscape, city,...) or weather conditions (sunny, cloudy, ...). Finally, coverage (i.e., ratio of false negatives) must be lower than 20\% in order to prevent false alarms and the unnecessary triggering of the interception action  (for instance). 

In order to reach a high level of reliability and availability, we choose to follow the recommendations promoted in the aeronautics domain, and specifically the guidance \cite{ConceptPaper}  released by the the EASA (European Union Aviation Safety Agency) and the soon to be published ED 324/ARP 6983 recommendations currently being developed by the SAE G34/EUROCAE WG114 working group~\cite{wg114}.
Another aeronautical standard, named the SORA \cite{habibi2023sora}, has been published to regulate drone flight to ensure safe operations in air traffic and environments. However, it does not address the integration of ML models into safety-critical systems. Applying the SORA could be considered in a further step to integrate an ML-based detection function in an interceptor drone (which is a better way to treat intrusion in sensitive area rather than an on-ground system).
\subsection{ED 324/ARP 6983} \label{sec:arp6983}

\begin{figure*}[hbtp]
    \centering
    \includegraphics[width=\textwidth]{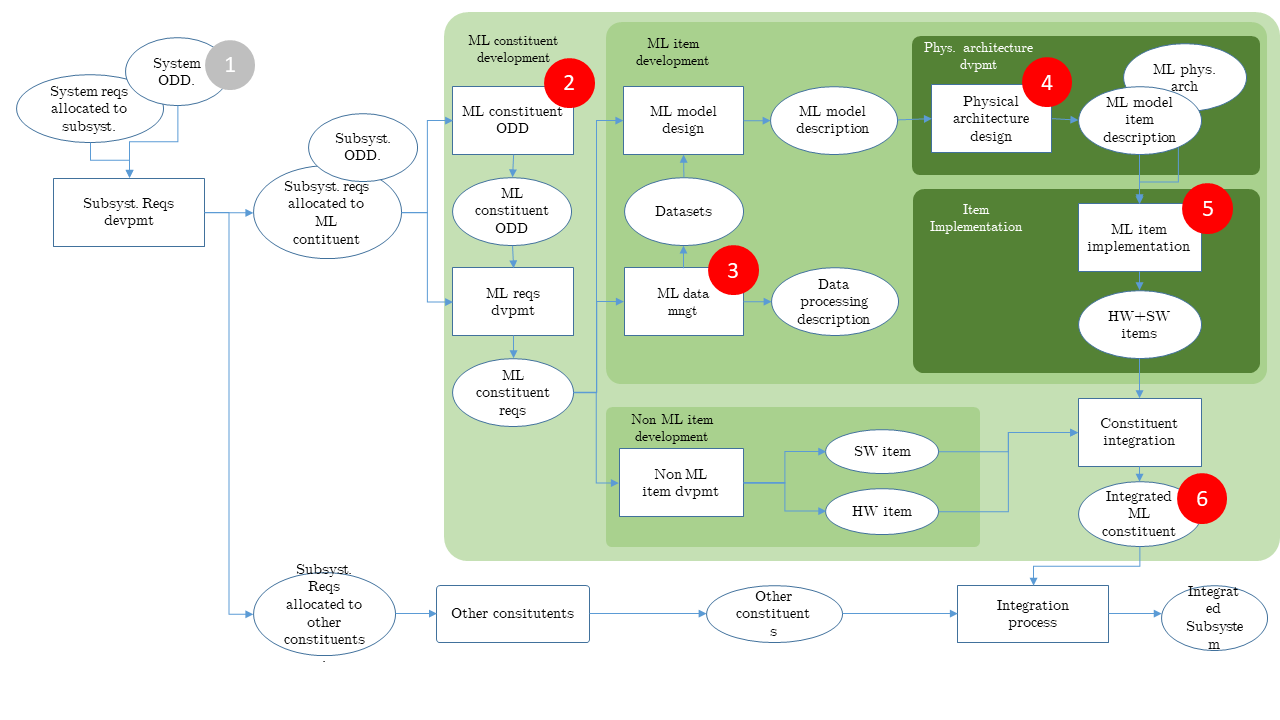}
    \caption{ARP~6983 simplified development workflow (adapted from \cite{gabreau:hal-03761946})} 
    \label{fig:arp6983} 
\end{figure*}

The ARP~6983 is a Process Standard for Development and Certification/Approval of Aeronautical Safety-Related Products Implementing AI\footnote{See \url{https://www.sae.org/standards/content/arp6983/}}. As of the date of redaction of this article, this document is still a work-in-progress, and a first public version is expected in Q2 2025. The ARP~6983 provides guidance that can be used as means of compliance for embedded AI. It complements existing practices to cover the specific issues raised by the introduction of AI/ML. Insights on the expected contents of the standard can be found in \cite{gabreau:hal-03761946}.

The ARP~6983 covers a significant part of the engineering activities for a AI/ML system, from the system/subsystem level down to the hardware and software items levels, through the ML constituent level\footnote{A ML constituent is defined in the certification guideline as a constituent containing the ML model(s) and its associated data processing.}.
An overview of the overall process is given on Figure~\ref{fig:arp6983}. Note that this diagram refers to an interim version of the standard that still may change before the official release of the standard is published. 

In this paper, the focus is placed on the activities of the standard labelled with a red tag (those in grey are mentioned for completeness). More precisely: 
the system Operational Design Domain (\ding{192}, §\ref{sec:sysodd}) and the other subsystem requirements are used to develop the ODD (Operational Design Domain) of the ML Constituent, or MLCODD (\ding{193}, §\ref{sec:mlcodd}) and, finally, the input dataset (\ding{194}, §\ref{sec:dataug}). 
The ML model designed to comply with the ML constituent requirements (\ding{195}) is refined into one or several ML Model Item Description(s) (MLMID, \ding{196}), which are implemented and deployed on the target hardware (\ding{197}, §\ref{sec:modeldesign}). 

\subsection{Contributions}

Our main contribution is the description of a partial process compliant with ARP 6983 applied in the development of an actual industrial system.
The main phases of the development process addressed in the paper are the following:  
\begin{itemize}
\item in phase~1, the system-level Operational Design Domain is specified and \emph{propagated} to the ML \emph{constituent} of the system,
\item in phase~2, the dataset is built (selected, augmented) in compliance with the ODD and the MLCOOD,
\item in phase~3, the ML model is designed (selected, adapted) so as to comply with the functional (ML performance) and non functional (memory footprint and latency) requirements, 
\item finally, in phase~4, different implementation paths are investigated considering traceability and latency concerns.
\end{itemize}

In addition from applying this process, we also propose the following technical contributions:
\begin{itemize}
    \item Biases on the input dataset have been identified and corrected by data augmentation;
    \item Compliance with detection performance and inference latency requirements has been addressed by conjointly (1)~improving detection performances by (1.a)~preventing information loss due to image redimensioning, (1.b)~achieving good sub-image overlap, and 
    (2) improving implementation efficiency thanks to (2.a) a quantized representation of the  ML model (FP16 and INT16) and (2.b) an efficient implementation of the GEMM matrix multiplication operator. 
\end{itemize}

There is a significant and increasing number of publications addressing the usage of ML in safety critical systems. When it comes to certification aspects, we can for instance mention the work on the ACAS-Xu in~\cite{damour-2021}. However, this work considers a very specific problem with a narrow operational domain (5 scalars).
More recently, papers address the certification vision-based landing such as \cite{denney23}. But none of them covers the full spectrum as we do and none addresses the drone intrusion detection problem. In that sense, our case study is representative of a (new) broader class of problems for which ML is considered useful.

The paper is organized as follows: 
Section~\ref{sec:odd} presents our approach to define the Operational Design Domain of our ML constituent; Section~\ref{sec:datadesign} describes the dataset design compliant to the ODD; Section~\ref{sec:modeldesign} and Section~\ref{sec:modeldeployment} describe respectively the model design and deployment process; Section~\ref{sec:conclusion} concludes the paper.

%% file: 2-ODD.tex
\section{ODD Specification}\label{sec:odd}
The first phase of our process is to define the Operational Design Domain (ODD). 
The ODD of a system is the allocation of the Operational Domain (OD) requirements to the system, the OD being a ``specification of all foreseeable operating conditions under which an end-product is expected (and should be designed) to fulfill its missions''~\cite{kaakai2023datacentric}. The ODD is a crucial element for the development of any ML-based system.  
Below, we first specify the ODD at system level and then refine it at ML constituent level, where the ML constituent only includes the ML model and its associated processing. 

\subsection{System ODD}\label{sec:sysodd}
The system ODD has been developed from a set of operational scenarios provided by domain experts. 
This set is deemed to cover the complete range of conditions in which the system must operate. 
For defining the operational scenarios, we use some terminology of the ISO 34503 \cite{iso34503},
a standard that proposes some concepts and requirements to enable the definition of an ODD of an automated driving system.
In particular,
an operational scenario determines 
(i) the environmental conditions,
(ii) the set of \emph{dynamic elements}, also referred to as \emph{objects} (intruder or drone, birds or other),
(iii) the set of \emph{scenery elements} (spatially fixed elements) of the system environment ("landscape", "sun", etc.) that must be considered and (iv) the set of attributes characterizing those (dynamic and scenery) elements (e.g., ``position in the sky'' for the ``sun'' element, ``type of drone'' for the ``intruder'' element, etc.). 
Here is an example of such a scenario.
\begin{scenario}
\it
is defined by:
\begin{itemize}
    \item \emph{Environmental conditions}: Time = 2 pm. Season = winter. Location: Europe. Atmosphere (nebulosity = none -- meaning that the weather is clear).
    \item
 \emph{Scenery elements:} The system is installed in an urban area with background buildings of high below 15m. 
 \item
\emph{Dynamic elements:} A 50cm x 50cm x 20cm drone arrives on the hand left side of the surveillance area (with orientation = (10°, 25°, 3°)) 
at a distance of 450m from the system, moving with a straight trajectory, in the direction of the system, at a constant speed of 1m/s.
Sun is visible (on the left hand side of the image).
\end{itemize}
\end{scenario}

The set of operational scenarios encompass many \emph{elements} and associated \emph{attributes} that are translated as a set of constraints 
that in fine defines the ODD. 
Those constraints can be numerous, highly complex or non tractable by human.
It may that an additional simplification step is necessary to 
aggregate some constraints into simpler ones, that can be more easily interpreted at the image level. For instance, the ``position of the sky'' and ``presence of clouds'' could be folded into a single attribute called ``lightning conditions''. 
In addition, the experts should define the \emph{realistic} distribution of elements (together with their attributes).
For the drone intrusion detection, some of the constraints obtained to define the ODD are:
\begin{itemize}
    \item The \textit{type of intruders} is in \{Quadrotor, Birotor\};
    \item The \textit{size of intruders} is within [0.5m, 1m];
    \item The \textit{type of area} is in \{urban, sub-urban, country side\};
    \item The \textit{time of the day} is in within [6am, 10pm];
    \item The \textit{lighting condition} is the range [sunny, slightly cloudy];
    \item etc.
\end{itemize}

To complete the ODD, the EASA guidelines \cite{ConceptPaper} \cite{kaakai2023datacentric} (\emph{Anticipated MOC DM-01-1})
and the ARP require to
identify particular conditions that need to be specified explicitly 1) so as to be surely taken into account (edge-cases) during training and testing;
or 2) on the contrary to be removed from the ODD (outliers). 
Those edge-case conditions may refer, for instance, to specific relations between attributes of the environment elements. The condition where an object has the same color as some other elements of the environment 
(e.g. painted intruder in green to reproduce the grass and hide easily)
is such an example in our context. Edge-case conditions could be generated by randomly sampling values of the various attributes identified in the ODD, but with a very low probability.

For example, the ODD given above allows some other objects such as distant airplanes or helicopters to appear in the surveillance area where they would be hard to distinguish from a drone (to some extent, some drones may be seen as a ``small airplanes''). These situations are considered rare and as outliers. Actually, they can be made as rare as necessary by forbidding to install the system in areas close to airports, for instance. The ODD must clearly address those outlier situations.

Finally, let us remind that the definition of the ODD is an iterative process. 
For instance, the latter constraints on edge-cases and outliers have to be ``reinjected'' in the ODD to complete it.
The model design could also lead to refine the ODD.

\subsection{ML Constituent ODD}\label{sec:mlcodd}
\begin{figure}[hbt]
    \centering
    \includegraphics[width=1\linewidth]{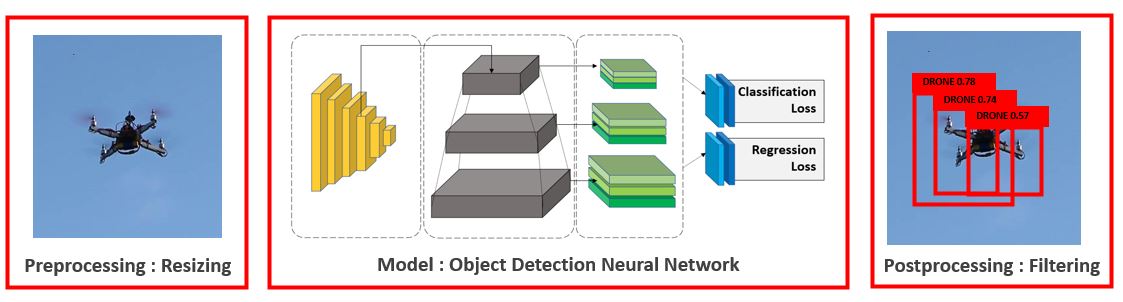}
    \caption{ML Constituent}
    \label{fig:ML Constituent}
\end{figure}

According to~\cite{wg114}, the ML constituent is the ``defined and bounded set of either hardware item(s) and/or software item(s) that implement ML''.  
In our case, the ML constituent is a software component (running on some piece of hardware) that takes as input images provided from a camera and generates as outputs data representing bounding boxes of objects detected in the image along with their classification. 
The ML constituent, figure \ref{fig:ML Constituent}, contains three main software components (the pre/post-processing and the ML model implementation).
The pre-processing is in charge of translating the raw image into a format expected by the ML model (e.g. resize high resolution images into 640x640).
Typical object detection algorithms are presented in \cite{objectdetection}.
The post-processing is in charge of computing the localisation of the intruder (if any) in the image and providing the absolute position on the area.

The ML constituent ODD (MLCODD) specifies ``the foreseeable operating conditions under which an ML Constituent is expected to work''~\cite{wg114}.  
In this paper, focus is placed on the image processing elements, so the MLCODD of interest refers to constraints related to the processed images. Some of those constraints  represent the ``projection'' of the system-level ODD constraints to the image domain; some others are related to the technical solutions used to implement the ML components. Examples are:
\begin{itemize}
    \item The \textit{size of objects} is in the range [20 px$^2$, 400 px$^2$]
    with 95\% of them in [20 px$^2$, 100 px$^2$];
    \item The \textit{position of objects in the image} is uniformly distributed in the 2 (geometrical) axes of the image;
    \item The \textit{main frequency components of the image}\footnote{\url{https://en.wikipedia.org/wiki/Frequency_domain}}
    is in the wavelength greater than 20 px (covering a range from  e.g., a solid clear sky background to a complex grass background);
    \item The \textit{mean brightness of the image} computed as the mean of the V value for the image coded in HSV (Hue Saturation Value) should belong to a pre-defined range;
    \item etc.
\end{itemize}

%% file: 3-Dataset.tex
\section{Dataset design}\label{sec:datadesign}
The second phase of our process is to create the dataset that is used during the training, validation and test phases. 
The dataset is built so as to comply with the definition of the MLCODD (see section \ref{sec:mlcodd}). In our case, we exploit private and public existing data sources including~\cite{fujii-2021,zhao20212nd}. 

\subsection{Biases analysis}
For the system to behave with the expected level of performance in operations, the dataset used during the training phase must reflect the distribution of situations that will be actually encountered during operations.  
These distributions and constraints are defined by the ODD.
Let us take as an example the position of the objects in the image.
The ODD states that their position is uniformly distributed on the camera image plane\footnote{As stated earlier, the system is also fitted with a radar that would normally place the camera axis in the direction of the object. Here, we consider that this feature is not reliable and ignore it.}

\begin{figure}[hbt]
    \centering
     \includegraphics[width=1\linewidth]{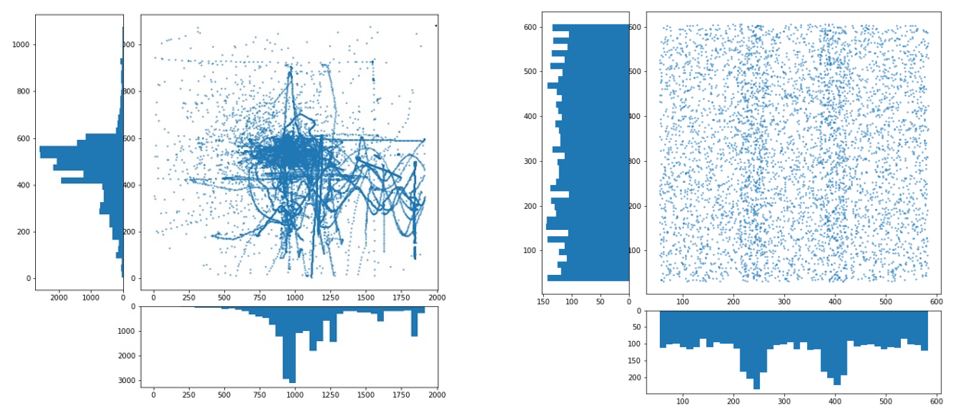}
    \caption{Object positions in the original dataset on the left and the augmented dataset with unbiased position on the right}       
    \label{fig:obj-pos}
\end{figure}

 Figure~\ref{fig:obj-pos} left hand side shows the spatial distribution of the objects in our initial dataset: 70\% of the objects are located in a very narrow area centered in the image. This is clearly not representative of the actual operational conditions (and thus not compliant with the MLCODD), for it would mean that most objects would fly towards the system from a far distance and remain centered on the camera axis. This is clearly a bias that may have a significant negative impact on the capability of the system to detect objects in 
 operating conditions such as
 the early stage of intrusion (i.e, when the objects enter the surveillance area from the side of the observation cone)
 and when the drone tries to leave the camera's view. 
 Having such a biased dataset during the learning phase would potentially lead the network to \textit{erroneously} correlate the presence of a target to its position in the image. This would also significantly reduce the detection accuracy for objects located at the edge of the image. 

\begin{figure}[hbt]
    \centering
     \includegraphics[width=0.5\linewidth]{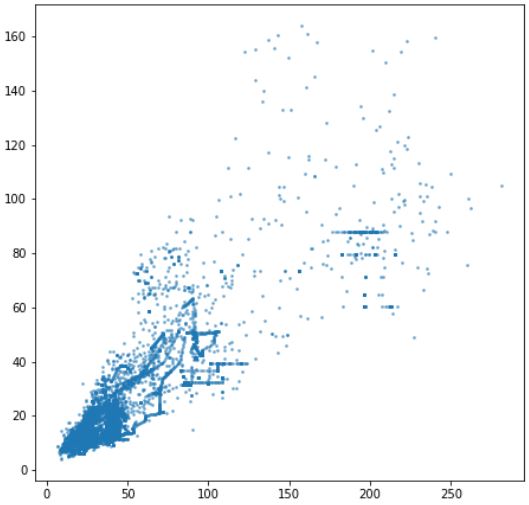}
    \caption{Distribution of size of object in the dataset}
    \label{fig:hist2D}       
\end{figure}

By displaying the size of the bounding boxes (Figure~\ref{fig:hist2D}), we observe that the dataset is compliant with the ODD with respect to object sizes. Indeed, almost 95 \% of the boxes are smaller than 100x100 pixels.

\subsection{Dataset augmentation}\label{sec:dataug}
In order to remove the biases and improve the representativeness of the dataset with respect to the MLCODD, 
the dataset must be enriched. There are several strategies to enrich the dataset among which making real data collection in real environment.
A real data collection is always a challenge because it is expansive and it can hardly reach the quantity / distribution / independence requirements.
For example, ensuring a large variety of intruders would imply to have access to many drones and program them to make several types of intrusion.
Collecting several images of one drone intrusion corresponds to one intrusion and dependent data.
Moreover, reaching a uniform drone position on the camera would be highly challenging.
This is the reason why, augmentation with image processing is often used, in addition to real data collection, to reach compliance with the ODD.
We have thus applied the following data augmentation techniques: 
\begin{itemize}
    \item generating images with objects at various positions and with various sizes;
    \item generating inlaying objects in various backgrounds (e.g., sky or urban background);
    \item generating new versions of existing images with modified brightness; 
    \item generating images with various numbers of objects (from 0 to 4).
\end{itemize}


To unbiased the dataset, we develop the algorithm \ref{alg:SpatialConverage} that is detailed hereafter.
The idea of the algorithm is to create new images from the original dataset by performing image transformations.
A possible transformation consists in transforming an 1080x1920 image with a drone at coordinate (486,921) (i.e. in the center)
into an 640x640 image with the drone at coordinate (486,510) (i.e. bottom right).
The spatial distribution of the new unbiased dataset with respect to the object position 
is shown on the right hand side of Figure \ref{fig:obj-pos}.
Thanks to data augmentation, the objects are uniformly 
distributed in the 2-axes.


\begin{algorithm}[hbt]
    \caption{Spatial Coverage Algorithm}
    \label{algo:training}
    \begin{algorithmic}[1] 
        \STATE Let D be the set of all 1080x1920 images 
        \STATE Let s be the size of the target image
            \FOR{$i : Image$ \textbf{in} D}
                \STATE Get the coordinate of the object in $i$
                \STATE Collate $i$ with 7 background images along each side of $i$.
                \STATE Get the coordinate of the object in the new reference frame
                \STATE Generate a random position of a window where the object is present and then crop the window
            \ENDFOR
        \STATE Save all new pictures in the dataset P with 640x640 images \end{algorithmic}
\label{alg:SpatialConverage}        
\end{algorithm}

\begin{figure}[hbt]
    \centering
    \includegraphics[width=.8\linewidth]{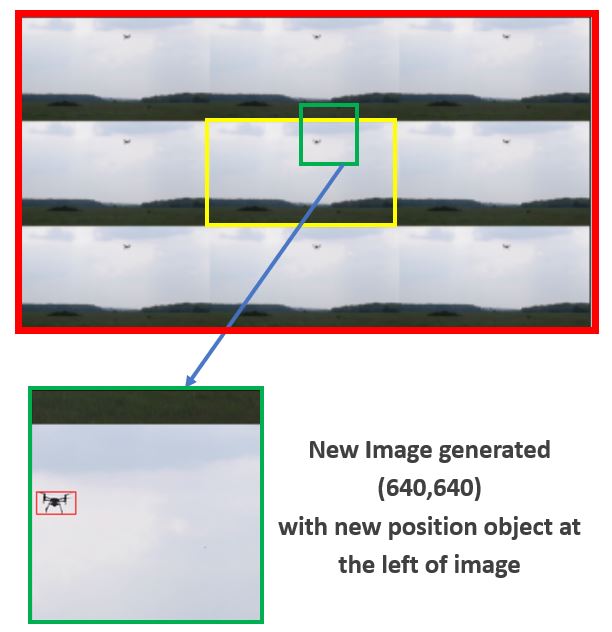}
    \caption{Algorithm \ref{alg:SpatialConverage} in action}
    \label{fig:ISpatial}
\end{figure}

The algorithm
\ref{alg:SpatialConverage} allows to generate new images with a specific size and a new object position.
Let us explain the algorithm via an example shown in the Figure~\ref{fig:ISpatial}.
First an image is picked (yellow in the picture), then is replicated nine times to produce a larger image (red). The yellow image can contain an intruder or can be a background.
A window (green) with a size of (640,640) is randomly selected and cropped. The resulted image  (in green on the bottom) is added to the dataset $P$. The algorithm does not resize the objects and allows to off-center them.

\begin{figure}[hbt]
    \centering
    \includegraphics[width=0.5\linewidth]{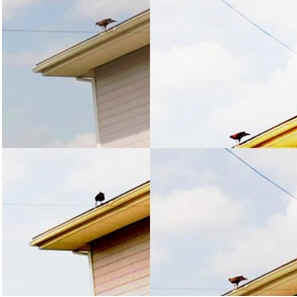}
    \caption{Mosaic dataset creation}
    \label{fig:Mosaic dataset}
\end{figure}

In addition to algorithm \ref{alg:SpatialConverage},
we create a ``mosaic dataset'' to increase the number of samples with different contexts. Figure~\ref{fig:Mosaic dataset} shows a mosaic image of size (640,640) which is an aggregation of 4 samples of size (320,320) with slight variations of brightness and image geometry.
This allows to address the brightness and main frequency components constraints of the MLCOOD.

\subsection{Compliance with the MLCODD}
\label{sec:bilan-MLCOOD}
We have analysed the initial dataset $D$ along the dimensions (e.g. size of object, position in the image, number of objects ...) identified by the MLCODD and compared with the constraints set by the MLCOOD.
As several constraints were not satisfied,
we have extended $D$ so as to comply with the definition of the MLCODD. 
In this paper, we illustrate some of those dimensions (at the end of section \ref{sec:mlcodd}) and provide the algorithm \ref{alg:SpatialConverage} to illustrate some image transformations to reach compliance.
In particular, we have unbiased the dataset with respect to the objects' positions by adding images with objects relocated at random places.
We have also increased
1) the coverage of drone attitude thanks to rotation and symmetry transformations (in the mosaic approach);
2) the representativeness of contexts with a large diversity of backgrounds and variation of brightness;
3) the situations to have from 0 to 4 intruders on images.
The final dataset $P$ is more than three times bigger than the initial one $D$.

The last step of the data management process is the splitting of the dataset into three subsets: training, validation and test datasets, each compliant with the MLCOOD.


%% file: 4-model.tex
\section{Model Design}~\label{sec:modeldesign}
The third phase of our process is to select and adapt a detection and localization algorithm in order for the ML constituent to perform the intended function. 
The model design was done following state-of-the-art machine learning approaches and an ad hoc tiling strategy to optimize the accuracy.

\subsection{Requirements}
\label{subsec:req}
The ML constituent, and the ML model, are designed to
realise the intended function (detect safely and quickly intruders in the sensitive area) in all operational scenarios defined during the ODD design.
In addition to the (MLC)ODD definition, the design phase has also identified requirements to be fulfilled.
There are several types of requirements including, but not restricted to, \emph{functional performance}, \emph{output format} compatible with the intended function
and \emph{real-time performances}.
\begin{requirement}[Functional Performance]
\it
The first type of requirements concerns the detection capacity of the model. 
    \begin{itemize}
    \item The ML model shall classify objects with an accuracy greater or equal to 90\% in areas A1 and A2, and greater or equal to 80\% in area A3.
    \item The ML model must have a false alarm rate of less than 20\%.     
    \item The ML model must have a non missed UAV rate of less than 20\%.     
    \end{itemize}
\end{requirement}
The ML constituent is expected 
to output the image coordinate 
of the intruder(s) if any, to provide the classification, to crop of the detected intruder(s)
and to compute their absolute position. All of these information are used for the decision-making (e.g. interception).
\begin{requirement}[Output format]
\it
 The ML model shall localize object in the image with a bounding box and 
 should output the bounding box coordinates, the classification and the confidence level.
\end{requirement}
The last type of requirements applies to the deployment and implementation of the ML model on the target.
\begin{requirement}[Real-time performances]
\it
The ML model shall be deployable on an Nvidia Xavier AGX target with the minimal use of COTS software and libraries.
Moreover, the ML model shall detect and localize objects in at most 50 ms.
\end{requirement}
\subsection{Model selection}
Object detection, classification, and localization tasks are usually done using deep learning neural networks. We experimented two different models: a one-stage model (YOLOv3~\cite{redmon2018yolov3}) and a two-stage model (Faster RCNN~\cite{girshick-2015}). While the accuracy was similar for both models, the inference latency of the two-stage model was incompatible with the \emph{real-time performance} requirements. In addition, the YOLOv3 model has the capability to detect objects at different scales, and this maps nicely to the different areas considered in the system. We choose the YoloV3 tiny \cite{inproceedings} model which is a refinement of the lighter YoloV3 model with optimized feature scaling, making it more efficient in terms of latency.


Another important design choice is to rely on a smart pre-processing.
During the design of the model, focus was placed on improving the detection of small objects, while keeping the architecture of the YOLOv3 component. To achieve this goal, we choose not to rescale the 1920x1080 camera image in order to preserve its information content. Instead, we decompose the high resolution image into several 640x640 tiles with some overlap, as described in the next paragraph and shown on Figure~\ref{fig:TilingStrat}. This solution reveals to be a good trade-off between detection performance, latency, and memory footprint. 

\subsection{Tiling Strategy}
\begin{figure}[hbtp]
    \centering
    \includegraphics[scale=0.4]{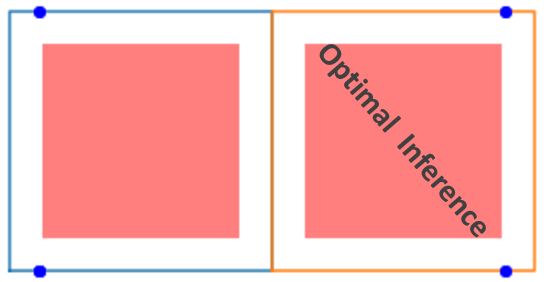}
    \caption{Optimal Inference Area}
    \label{fig:OptimalInference}
    \vspace*{-0.3cm} 
\end{figure}

The transformation of a high resolution 1920x1080 image into multiple smaller images (or \emph{tiles}) without image compression can be done in several ways.
Such tiling strategy depends on the tile size (here fixed at 640x640)
and  the targeted overlap between tiles \cite{9025422}. 
Each tile must then be analyzed by the YoLo model leading to several model inferences to cover the full high resolution image. This has a direct impact on the real-time performances (and the latency).
In terms of implementation latency, having no overlap 
is the best solutions since there will be less tiles. 
However, in that case, 
the detection algorithm  only gets a partial view of objects located at the boundaries in each tile. 
Moreover, models are known to badly detect objects (even if complete) on boundaries due to phenomena of blind spots \cite{alsallakh2020mind} and spatial bias \cite{zheng2023zone}.
So we must choose the size of the overlap to force the intruder(s) to be in at least one optimal inference area, as shown on Figure~\ref{fig:OptimalInference}.

\begin{figure}[hbtp]
    \centering
    \includegraphics[scale=0.4]{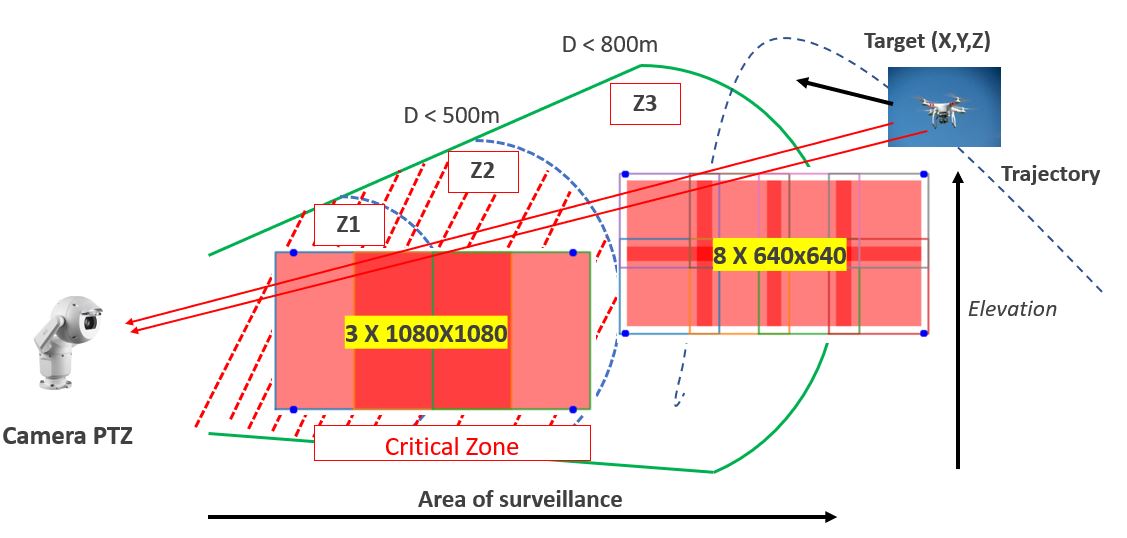}
    \caption{Tiling Strategy according to the Area}
    \label{fig:TilingStrat}
\end{figure}

As a consequence, 
the objective of our tiling strategy is to find an \emph{optimal decomposition trade-off} that allows some overlap for the detection accuracy and 
that produces reasonable number of model inferences  for the detection latency.
We have defined two strategies depending on the detection areas since they do not share the same functional and real-time requirements.
Indeed, in areas 1 and 2, the drone size in pixel is larger than in area 3 (thus the detection performance is easier) but the latency is shorter.
We thus select for those areas the configuration of Figure~\ref{fig:TilingStrat} left hand side composed of three 1080x1080 tiles with an overlap of 50\%.
The 4 blue points represent the input space to cover. 
Because the YoLo model expects 640x640 tiles, a resize processing is applied on each  1080x1080 tile.
Since the drone size is large, the resizing will not reduce it too much and the YoLo performance remains in the range of acceptable models.
The overlapping is important (50\%) to increase the accuracy on the frontier between the left and right parts of the image.
Finally, having 3 tiles leads to a reduced latency.

The case of area 3 is different since the drones are far away and their size in pixel is small. 
The  tiling scheme for this area is shown on Figure~\ref{fig:TilingStrat} right hand side. It is composed of eight 640x640 tiles with an overlap of 30\%. 
Resizing would degrade too much the performance thus the input image is just decomposed. Having more than 8 tiles is challenging for the implementation and the latency constraints.
The two tiling schemes, summarized in Table~\ref{tab:DWBencharea}, meet latency and memory requirements and provide similar object detection performance for each area.  

\begin{table}[hbtp]
    \centering
    \begin{tabular}{lcc}
    \hline
    AREA  & A1 and A2 & A3   \\ \hline \hline
    Tiling & (3,1080,1080)    & (8,640,640)     \\ \hline 
    Overlap & 50\%   & 30\%      \\ \hline     
    Resizing & 56\% on one axis  & None     \\ \hline     
    \end{tabular}
    \caption{Tiling configuration for inference optimization}
    \label{tab:DWBencharea}
\end{table}

\subsection{Model optimization}


In order to lower the computational footprint of the YOLOv3 algorithm while maintaining the detection performances, we modified its backbone by replacing the 7 2D-convolution layers with depth-wise separable convolution (DSC) layers. The use of DSC dramatically reduces the number of operations and the memory footprint. Such layer is in particular used in the MobileNet model to support efficient object detection on embedded devices~\cite{howard2017}.



%% file: 5-deployment.tex
\section{Model implementation and deployment }\label{sec:modeldeployment}
The last phase of our process is to implement and deploy the design model. 
The implementation must ensure that all the requirements listed in section \ref{subsec:req} are satisfied. In addition, we also consider ARP 9683 development assurance concerns, including the capability to demonstrate traceability between the ML model and its implementation.

\subsection{Implementation approach}
\begin{figure}[hbt]
    \centering
    \includegraphics[width=0.4\textwidth]{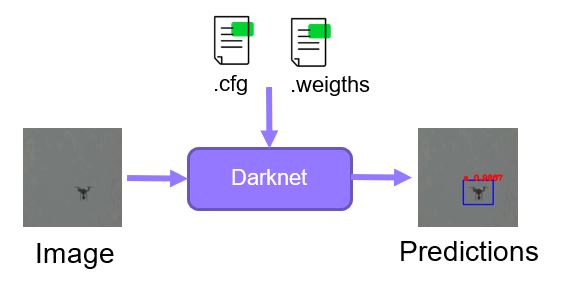}
    \caption{Darknet Framework}
    \label{fig:DK-FP}
\end{figure}

The target as already mentioned is the NVIDIA Xavier AGX platform that comes with 8 Carmel ARM-core, a GPU and a NVDLA (NVIDIA Deep Learning Accelerator). 
We selected the Darknet implementation 
framework~\cite{redmon2018yolov3} that supports many YoLo
object detection and classification algorithms, including the YoLov3 tiny that we selected. Darknet is open-source, which means that it can be analysed and possibly assessed for traceability and semantic preservation analysis. It has two back-ends: C code for CPU and CUDA code for NVIDIA GPU targets.
The training of the model was done with the Keras framework.
We then manually describe the model architecture of Keras in a ".cfg"  textual file (layers and operators) and export the parameters of each layer in a ".weights" binary file (coded in the IEEE754 format and little endianness). 
These two files must be consistent with each other and correspond to the ML model description (MLMD) provided at the end of the model design. 
These files are loaded as input by Darknet to generate the (C or CUDA) code, see figure \ref{fig:DK-FP}.

\subsection{C code generation for CPU}
We first investigate a CPU-only implementation and port the executable on one ARM code of the Xavier.
The generated C code is similar to the neural network description which facilitates traceability activities.
Several tests were made and the behaviour on the target was similar to one observed in the learning framework Keras.
Unfortunately, the real-time requirements are not satisfied.

\begin{figure}[hbtp]
    \centering
    \includegraphics[width=0.45\textwidth]{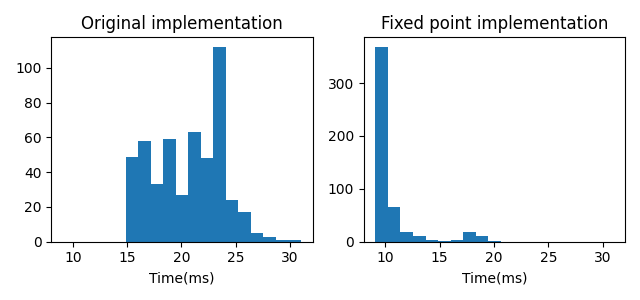}
    \caption{Original Darknet GEMM implementation \textit{vs.} fixed point modified version (latency for layer 24)}
    \label{fig:orig-FP}
\end{figure}

In order to compensate the limited performances of the CPU, we investigate the use of fixed-point arithmetic, on particular on the convolution layers.
Darknet generates Generalized Matrix Multiply (GEMM) \cite{chellapilla:inria-00112631} based implementation.
We optimized GEMM operator with ideas inspired by~\cite{geyer2024efficient,KhalifaM21,goyal2021fixedpoint}. 
\autoref{fig:orig-FP} shows the latency figures for layer 24 of the YOLOv3 model, using 32-bit floating point arithmetic (left side) and 16-bit fixed point arithmetic (right side). We observe a 50\% latency drop between the two implementations, a significant reduction of the measurement dispersion, and a reduction of 29\% of the Observed Worst-case Execution Time (OWCET). Unfortunately, this performance level still does not meet our requirements. 

\subsection{Cuda code generation for GPU}
We then try the Cuda code generated by Darknet on the NVIDIA GPU. In this case, layer 24 of the FP32-encoded YOLOv3 model is executed in around 1.2ms, which represents around 7\% percent of the latency measured on the CPU. Another optimization was achieved by replacing the classical convolution operator used in the initial model by a 
depth-wise separable convolution (DSC)
(already discussed in Section~\ref{sec:modeldesign}). This optimization led to a reduced inference time for each tile of (640,640).
The \autoref{tab:DWBench} summarizes the latency on the NVIDIA Xavier AGX platform. 

\begin{table}[hbtp]
    \centering
    \begin{tabular}{lcc}
    \hline
    Model (FP32)  & YOLOv3 & YOLOv3 DSC    \\ \hline \hline
    Inference time & 26ms    & 20ms  \\ \hline 
    \end{tabular}
    \caption{YOLOv3 inference}
    \label{tab:DWBench}
\end{table}

Finally, a second level of optimization was done by modifying Darknet to generate Cuda code in half precision (i.e. FP16) to reduce memory footprint and latency. 
By applying FP16 quantization \textit{to the entire neural network}, the total inference time decreases to 15 ms, a 25 \% improvement over the initial FP32 model.
Note that we also benchmarked different batch schemes to process the 8 tiles of the full resolution image (see~\autoref{sec:modeldesign}). We did not gain any latency benefit with such an approach.

\subsection{Optimized Cuda code for GPU}
The CUDA FP32 and FP16 implementations meet the latency requirements, but they rely on the closed-source cuBLAS library\footnote{\url{https://docs.nvidia.com/cuda/cublas}.}.
Using COTS libraries in safety-critical systems can be discouraged because traceability analysis is difficult and static WCET analysis could be unattainable. This is the reason why, we tried our own GEMM operator implementation inspired by the article of\footnote{\url{https://siboehm.com/articles/22/CUDA-MMM}}. In particular, we applied some optimizations, taking into account the specific structure of the network and the GPU platform.

\begin{figure}[hbtp]
    \centering
    \includegraphics[width=0.45\textwidth]{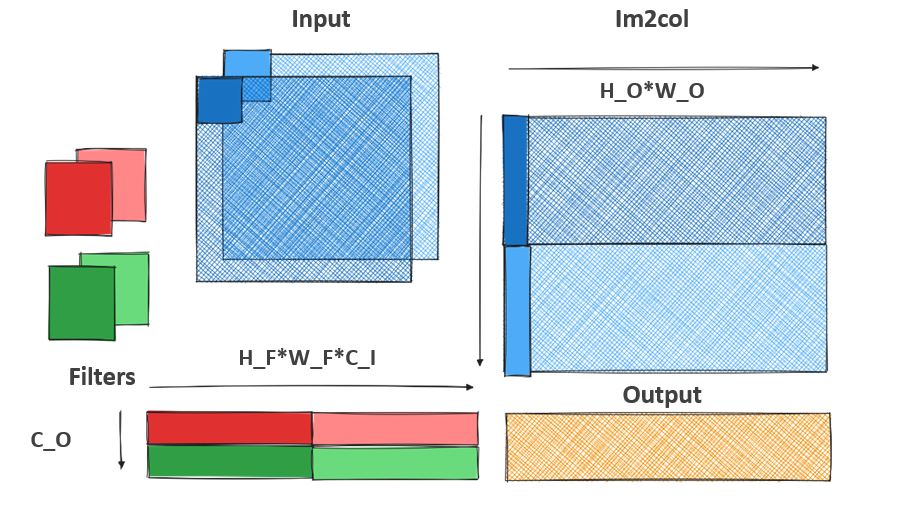}
    \caption{im2col tensor}
    \label{fig:im2col}
\end{figure}

We remind that to execute a convolution with a GEMM operator, the input 3D tensor is translated into a 2D matrix.
A 3D tensor is defined by its three dimensions (H,W,C)
where H refers to the height, W the width and C the number of channels.
The classical algorithm to translate a 3D tensor into a matrix for GEMM is the \texttt{im2col} method~\cite{chellapilla:inria-00112631}
shown in~\autoref{fig:im2col}. 
Note that nothing needs to be done for the kernel.

There is a huge variability in the tensor sizes that are exchanged within the layers of the YoLo architecture. 
Indeed, the first layers are made up of large images with few channels (large (H,W), small C), while the last layers are made up of small vignettes with a large channel depth (small (H,W), large C). 
We thus have two main configuration types: 
\begin{itemize}
\item C1 : First Layers : (Large (H,W), Small C)
\item C2 : Last  Layers : (Small (H,W), Large C)
\end{itemize}
These two types of tensor impact directly the size of the 2D matrix produced by the im2col function.
The CUDA GEMM operator then necessitates  different optimizations depending on the size of generated matrix.

\begin{figure}[hbtp]
    \centering
    \includegraphics[width=0.45\textwidth]{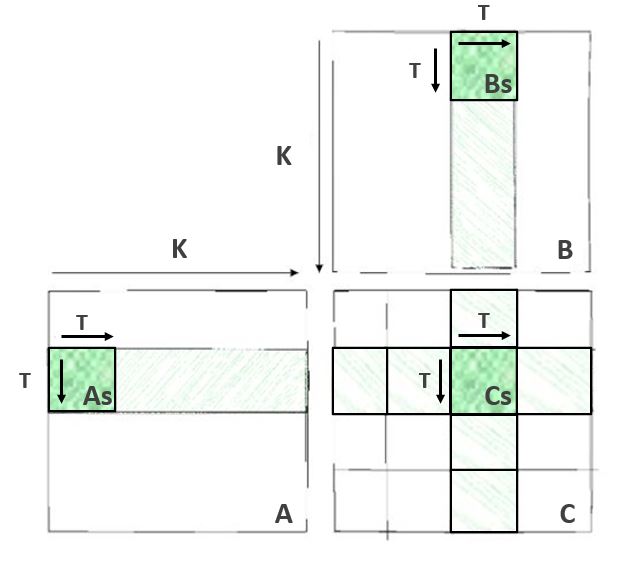}
    \caption{Matrix multiplication accumulation}
    \label{fig:MMA}
\end{figure}

To perform the  matrix multiplication $A \times B$, 
GEMM decomposes the matrices $A$, $B$ and $C$ into blocks $A_s$, $B_s$ and $C_s$.
Those blocks contain another level of decomposition (micro-panels and tiles) 
\cite{gemmblock,iryna24}
that we refer to here as chunk.
Each chunk
of the two matrices A and B is loaded in a shared memory array 
and a thread block on an SM calculates the element of the chunk of matrix C.
Note that the C chunk is partially computed and needs to be accumulated with several block operations. 
\autoref{fig:MMA} shows how the global matrix multiplication is decomposed using small blocks. The block parameters are: 
\begin{itemize}
\item M, N and K: dimensions of A (M,K) and B (K,N);
\item BM: Number of rows in blocks Cs and As;
\item BN: Number of columns in blocks Cs and Bs;
\item BK: Number of As columns and Bs rows;
\item TM: Number of elements of C computed by each thread.
\end{itemize}

A GPU is composed of multiple SM (Stream Multiprocessor)
where a SM is a general purpose processor (with cache, shared memory, etc.) able to execute several thread blocks in parallel. Each thread in a thread block is executed on the same SM. 

We propose to adapt the GEMM algorithm by adjusting the matrix block parameters as efficiently as possible according to the tensor configurations C1 and C2. 
To find the optimal chunk decomposition, we have
(i)~maximized the occupancy of the SM (we set the registers to 32 per SM to increase the number of 4 blocks used per thread to reach 100 \% of occupancy), (ii)~reduced the data movements on shared memory by optimizing the number of element calculated per thread so as to reduce the load of elements of As, Bs. In addition, the reproducibility of the results was also verified. All experiments are made in float32.

As shown in Figure \ref{fig:MMA}, matrices A and B are decomposed into blocks. These blocks ``traverse'' the rows of matrix A and the columns of matrix B to compute the blocks of the output matrix C.  In order to maximize performances, the size of these blocks is chosen so that they fit into shared memory. Then we optimize the thread in a thread block to load multiple elements in A and B to reduce data movement in shared memory. Moreover, the im2col operation provides a contiguous mapping of data in memory that enables  to optimize the calculus in the same block of threads.
In our case, the values of those parameters are given on \autoref{tab:DWBench}.

\begin{table}[hbtp]
    \centering
    \begin{tabular}{lcc}
    \hline
    GEMM Tiling & First Layer (C1) & Last Layer (C2) \\ \hline \hline
    M & 16          & 1024 \\ \hline 
    N & 409600      & 400 \\ \hline
    K & 27          & 4608 \\ \hline
    BN & 16         & 64 \\ \hline
    BM & 16         & 64 \\ \hline
    TM & 8          & 8 \\ \hline
    BK & BN/TM=2  & min(BN/BM)/TM=8 \\ \hline
    \end{tabular}
    \caption{GEMM block decomposition parameters}
    \label{tab:DWBench-gell}
\end{table}

\begin{figure}[hbtp]
    \centering
    \includegraphics[width=0.45\textwidth]{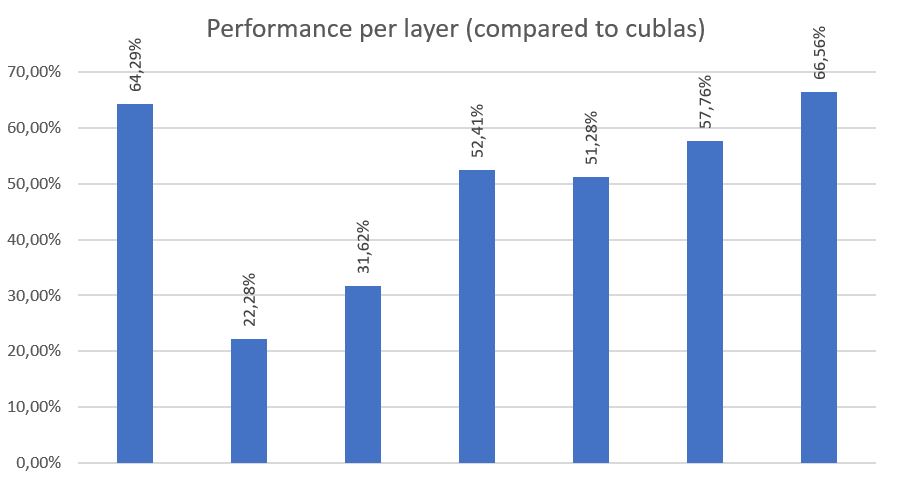}
    \caption{Performances of the GEMM implementation}
    \label{fig:gemm}
\end{figure}

Performances of our implementation with respect to CuBLAS' is given on~\autoref{fig:gemm}. The performance remains lower than the highly optimized version implemented in CuBLAS by NVIDIA's engineers, but it is sufficient to meet our latency requirement. 

%% file: 6-conclusion.tex
\section{Conclusion}\label{sec:conclusion}

The purpose was to design a surveillance system to detect and localize intrusions of UAVs in sensitive areas.
To support the design, we decided to follow part of the  ED 324/ARP 6983 guidance in order to help us increase the confidence and reliability of the design.
We have addressed, in a pragmatic way, 
some important issues raised when integrating a ML component in a system performing critical, real-time functions. 
We have proposed a partial process 
compliant with the ARP 6983
that helped us reaching a much higher quality / confidence in the system.
The process
 focuses on:
\begin{itemize}
    \item the ODD and MLCODD definition. This was possible with the definition of operational scenarios and their translation into constraints that can be mapped in the image domain;
\item the dataset design.
We have shown how biases on existing datasets have been detected and corrected via an appropriate data augmentation. 
The verification of the dataset compliance with the MLCOOD is done by providing insights of the dataset, such as the distribution of scenery elements with respect to the constraints identified for the ODD; 
\item the ML model design in order to reach the expected functional performances while integrating implementation constraints.
This results from a classical precision \textit{vs.} latency trade-off by preventing the loss of useful information thanks to the use of a tiling strategy instead a simple image resizing;
\item the implementation with optimizations.
Again, we addressed conjointly performance and dependability, by developing an ad-hoc implementation of the most demanding operator (GEMM) that allows to reach an acceptable performance level with a full traceability of the source code to the input model. 
\end{itemize}


As stated previously, we have only addressed a small subset of issues raised by embedding ML components.
In future work, we will extend our work in several directions including in particular (i)~a more precise and complete definition of the ODD in order to improve the quality of the datasets and, possibly, detect the out-of-ODD conditions in which the system cannot operate safely, (ii)~a more traceable implementation of the model (currently, traceability ends with the NVIDIA drivers and hardware comes into play).  

Finally, we will consider architectural means to alleviate the remaining and irreducible concerns raised by ML. In particular, we will consider the addition of system-level monitoring and mitigation means. Finding independent (e.g., non-ML) solutions remains a challenge considering that the function essentially relies on image processing. 

